\documentclass[a4paper]{article}

\usepackage{INTERSPEECH2019}
\usepackage{cite,url,float,balance,subcaption,color}
\title{Large-Scale Speaker Diarization of Radio Broadcast Archives}
\name{Emre Y\i lmaz$^1$, Adem Derinel$^1$, Zhou Kun$^1$, Henk van den Heuvel$^2$, \\ Niko Brummer$^3$, Haizhou Li$^1$, David A. van Leeuwen$^4$}
\address{
  $^1$ Dept. of Electrical and Computer Engineering, National University of Singapore, Singapore\\
  $^2$ CLS/CLST, Radboud University, Nijmegen, Netherlands\\
  $^3$ Cyberupt BV, South Africa\\
  $^4$ ICIS, Radboud University, Nijmegen, Netherlands}
\email{emre@nus.edu.sg}

\begin{document}

\maketitle
\begin{abstract}
This paper describes our initial efforts to build a large-scale speaker diarization (SD) and identification system on a recently digitized radio broadcast archive from the Netherlands which has more than 6500 audio tapes with 3000 hours of Frisian-Dutch speech recorded between 1950-2016. The employed large-scale diarization scheme involves two stages: (1) tape-level speaker diarization providing pseudo-speaker identities and (2) speaker linking to relate pseudo-speakers appearing in multiple tapes. Having access to the speaker models of several frequently appearing speakers from the previously collected FAME! speech corpus, we further perform speaker identification by linking these known speakers to the pseudo-speakers identified at the first stage. In this work, we present a recently created longitudinal and multilingual SD corpus designed for large-scale SD research and evaluate the performance of a new speaker linking system using x-vectors with PLDA to quantify cross-tape speaker similarity on this corpus. The performance of this speaker linking system is evaluated on a small subset of the archive which is manually annotated with speaker information. The speaker linking performance reported on this subset (53 hours) and the whole archive (3000 hours) is compared to quantify the impact of scaling up in the amount of speech data.
\end{abstract}

\noindent\textbf{Index Terms}: Speaker diarization, speaker linking, speaker and cluster impurities, longitudinal broadcast data, x-vectors
\vspace{-0.2cm}
\section{Introduction}
\vspace{-0.1cm}
Speaker diarization (SD) in the conventional sense has been achieved by clustering segments with similar speaker characteristics to reveal the number of speakers involved in a conversation and label who speaks when. Various top-down and bottom-up approaches have been proposed in which the diarization starts with small and large number of clusters respectively and iteratively converges to an optimum number of clusters~\cite{anguera2012}.

Large-scale SD techniques are employed on considerably larger amount data to perform SD in a computationally tractable manner. To deal with the computational restrictions of conventional SD techniques due to the major increase in speech data, multiple previous work proposed to perform SD in multiple stages~\cite{meignier2006,vanleeuwen2010,huijbergts2012,ferras2012,ferras2016}. The idea is to first perform standard SD on smaller units such as the tapes in a broadcast archive or recordings in a meeting archive and then link the speakers appearing in multiple times using a metric to assign a similarity score to all pseudo-speaker pairs which are hypothesized in the first stage. One common way of achieving the linking is to apply agglomerative clustering-based speaker clustering with different linking strategies such as single, Ward and complete-linkage \cite{ferras2016,ghaemmaghami2016}. In these previous works, the speaker clusters modeled as multivariate Gaussian mixture models (GMM) are merged until a certain stopping criterion is met.

More recently, SD techniques employing speaker clustering using probabilistic linear discriminant analysis (PLDA) for calculating similarity scores between clusters have been proposed \cite{prazak2011,sell2014,woubie2016,garciaromero2017}. In this setting, speech segments are represented in the form of i-vectors~\cite{dehak2011} and x-vectors~\cite{snyder2017,snyder2018}. As mentioned in multiple earlier work, e.g.~\cite{sell2014}, this SD framework provides better SD performance compared to the conventional GMM-based approaches using Bayesian information criterion~\cite{chen1998} to assign cluster similarity.

In the FAME! Project, we have developed a spoken document retrieval system for the radio broadcast archives of Omrop Frysl\^{a}n (Frisian Broadcast), the regional public broadcaster of the province Frysl\^{a}n in the Netherlands. This system relies on automatically generated transcriptions hypothesized by a code-switching automatic speech recognition system~\cite{yilmaz2018_1} and speaker labels generated by a modern speaker recognition system developed using the resources~\cite{yilmaz2017_3} with the ultimate goal of making these archives searchable. 

In this paper, we focus on the development of an experimental setup for speaker linking research on very large speech corpora and describe our initial efforts to build a large-scale SD system which would perform SD and linking in two stages. For this purpose, we created a new SD corpus, dubbed as the FAME! SD corpus, with 6494 digitized tapes from the radio broadcast archive that contains more than 3000 hours of Frisian-Dutch speech. The evaluation of the SD performance is done on the 82 subsegments that appear as a part of longer tapes. These subsegments were manually annotated with speaker information during the initial data collection described in \cite{yilmaz2016}. 

As a technical contribution, we introduce a new speaker linking approach that uses x-vectors with PLDA for assigning similarity scores to pseudo-speaker pairs. The tape-level SD performed in the first stage uses an off-the-shelf SD toolkit ensuring a reasonably accurate initial speaker clustering which are later used for the linking performed in the second stage. The speaker linking results provided by the proposed system on 82 partly annotated tapes and the whole archive are presented in terms of diarization error rate and speaker and cluster impurities. The longitudinal, multilingual and diversely accented nature of this speech archive introduces new challenges to modern large-scale SD systems which is yet to be explored in the future.
\vspace{-0.2cm}
\section{FAME! speaker diarization corpus}
\label{sec:database}
\vspace{-0.1cm}
The Frisian-Dutch speech data has been collected in the scope of the FAME! (Frisian Audio Mining Enterprise) Project with the aim of building a spoken document retrieval system\footnote{Available online at https://zoeken.fame.frl/} for the disclosure of the archives of Omrop Frysl\^{a}n (Frisian Broadcast) covering a large time span from 1950s to present and a wide variety of topics. Omrop Frysl\^{a}n is the regional public broadcaster of the province Frysl\^{a}n. It has a radio station and a TV channel both broadcasting in Frisian and is the main data provider of this project with a radio broadcast archive containing more than 3000 hours of recordings. The longitudinal and bilingual nature of the material enables to perform research on various research topics including speaker tracking and diarization over a large time period.
\begin{figure}[t]
\centering
  \includegraphics[trim=0cm 1.25cm 0cm 2.5cm, width=2.82in]{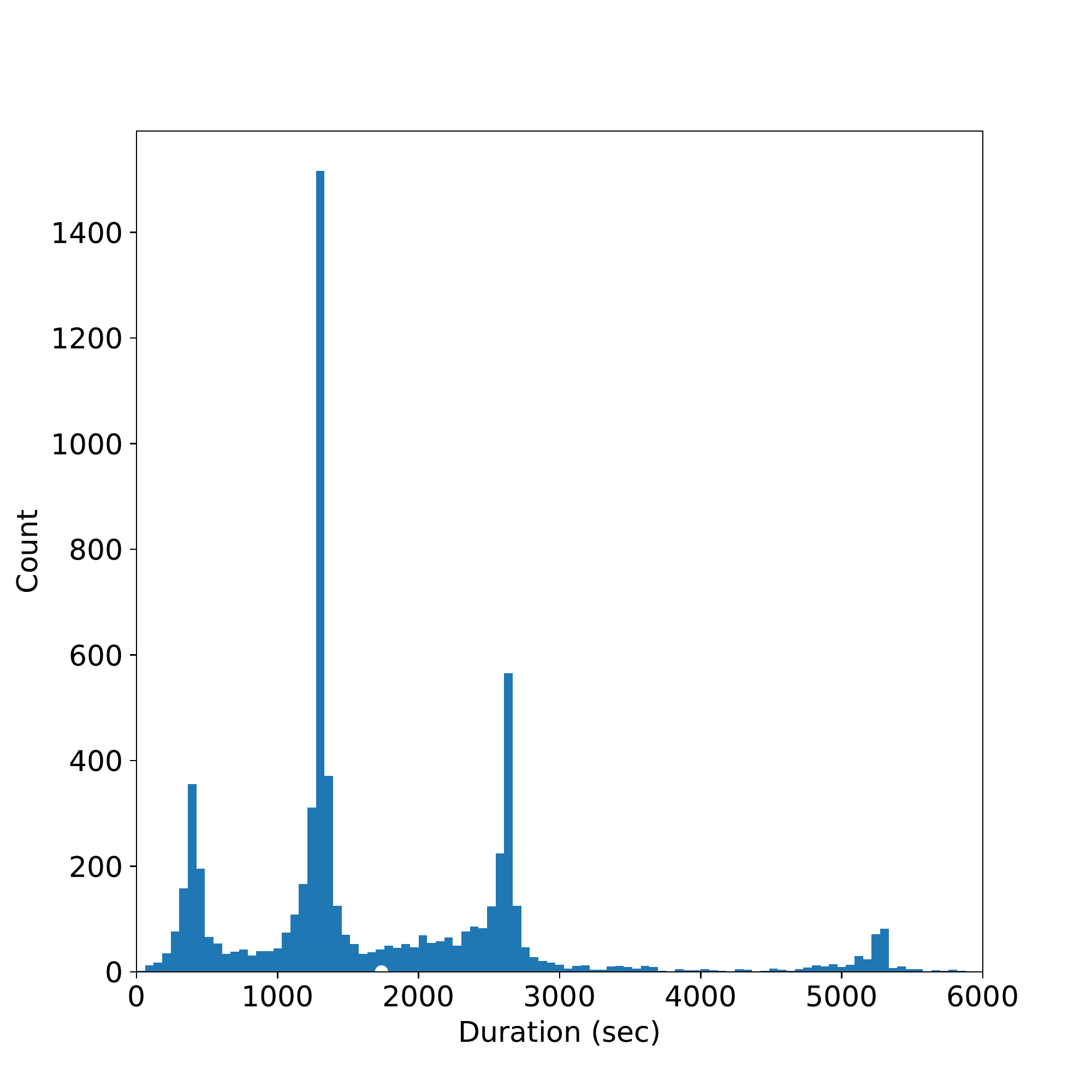}
  \caption{Duration distribution of the tapes in the radio archive}
  \vspace{-0.5cm}
  \label{fig:hist}
\end{figure}

A small subset of the radio broadcast recordings has been manually annotated by two native Frisian and Dutch speakers. The annotation protocol designed for this code-switching Frisian data includes the orthographic transcription containing the uttered words, speaker details such as the gender, dialect, name (if known) and spoken language information. To get more precise information about the speaker details, all available meta-information of the radio broadcasts is also used during the annotation. Further details of this corpus are given in \cite{yilmaz2016}. In the last years, we have created two publicly available speech corpora for CS ASR \cite{yilmaz2016_2} and speaker recognition \cite{yilmaz2017_3} research using the manually annotated and raw data extracted from these archives. After having the complete archive digitized last year, we have created a third speech corpus that is designed for large-scale SD research. Unlike previous corpora, this component will not be included in the open Frisian resources due to intellectual property rights.

The FAME! SD corpus contains 6494 tapes whose durations are distributed as shown in Figure~\ref{fig:hist}. The mean duration is 1737 seconds (28 minutes, 57 seconds) which is marked in white on the histogram. All available tapes that are provided by the broadcast after the digitization have been included except a few corrupted ones containing silence only. The evaluation of the speaker linking performance has been done on 82 partly annotated tapes. Each of these tapes has a 5-minute subsegment which has been annotated during the initial manual annotation effort \cite{yilmaz2016}. The exact positions of each annotated subsegment (start and end times) in the corresponding tape are manually determined and recorded in a text file which is included in the corpus. The total duration of these 82 tapes is 53 hours and the total duration of the annotated subsegments is 7 hours 20 minutes. There are 215 speakers in the annotated subsegments, 154 with known and 61 with unknown speaker name. According to the ground truth transcriptions, there are 22 annotated speakers appearing in more than one tape and 10 annotated speakers appearing in five or more tapes. The annotated subsegments are equally divided into a development and test set, each containing 41 subsegments with approximately 3 hours 40 minutes of speech. 
\vspace{-0.2cm}
\section{Large-scale speaker diarization}
\label{sec:sd}
\begin{figure*}[ht!]
  \centering
  \includegraphics[trim=0cm 0.5cm 0cm 2cm, width=5in]{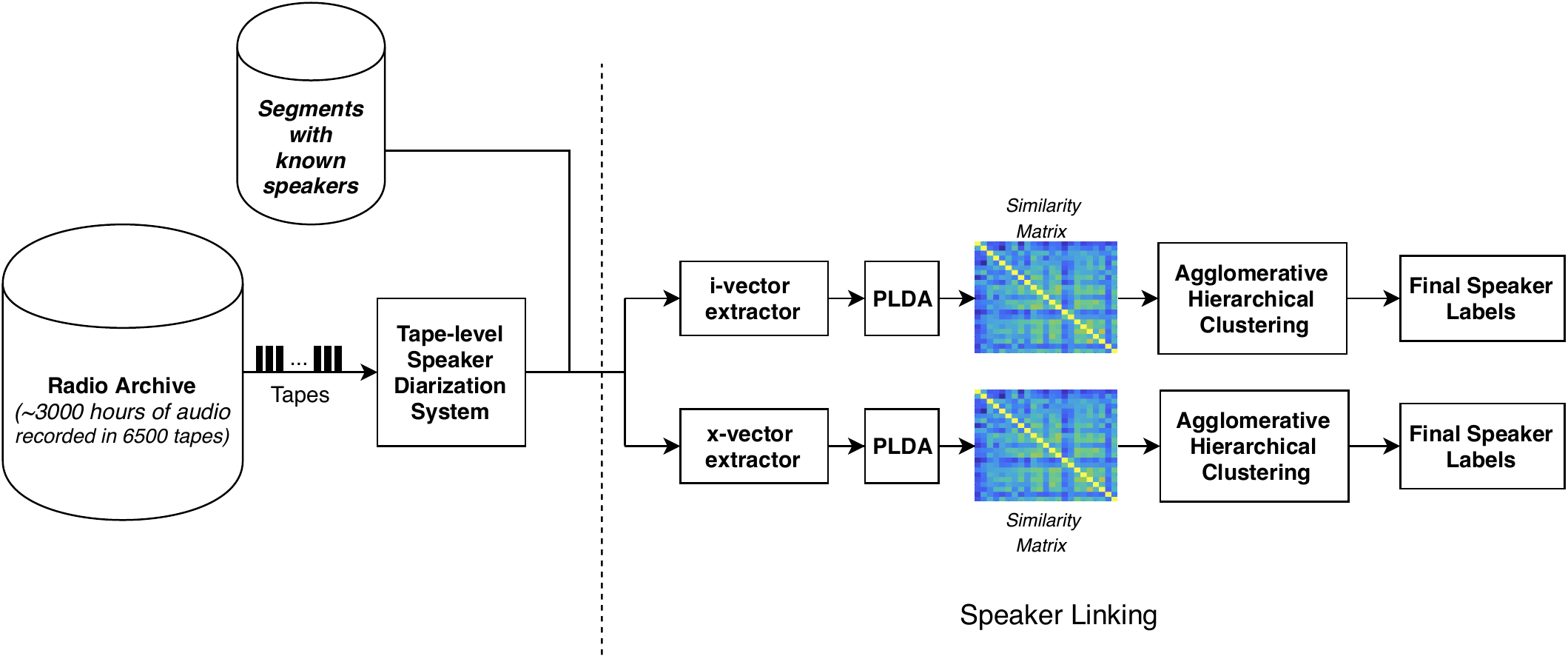}
  \caption{Block diagram of the large-scale speaker diarization and identification systems}
  \vspace{-0.5cm}
  \label{fig:overview}
\end{figure*}

\vspace{-0.1cm}
This section describes the details of the large-scale SD system applied to the radio archive described in Section \ref{sec:database}. As conventional approaches to SD cannot be employed in our setting due to intractable computational burden, we opt for a two-stage diarization scheme, which is illustrated in Figure~\ref{fig:overview} and detailed in the upcoming subsections. 

\vspace{-0.2cm}
\subsection{First stage: Tape-level diarization}
\label{ssec:tapelevelSD}
\vspace{-0.1cm}
The initial pseudo-speaker labels are obtained using the publicly available LIUM toolkit \cite{lium}. Focusing on the speaker linking component in the second stage, the goal in the first stage is to obtain a pseudo-speaker labeling of reasonable quality. The clustering parameters have been chosen to favor a mild overestimation of the actual number of speakers in the first stage to prevent irreversible merging of speakers on tape level. 

The incorporated SD system uses Gaussian models performs acoustic segmentation and clustering based on a Bayesian information criterion (BIC)~\cite{chen1998} followed by a GMM-based speaker clustering by maximizing cross likelihood ratio (CLR) \cite{reynolds1998} measure to obtain optimal speaker clusters. Further details of the LIUM toolkit can be found in \cite{lium}.
\vspace{-0.2cm}
\subsection{Second stage: Speaker linking and identification}
\label{ssec:sl}
\vspace{-0.1cm}
Linking and identifying pseudo-speakers is a straightforward step towards improving the quality of the automatically generated speaker labels. The goal is to assign identical labels to the speakers appearing in multiple tapes, e.g. presenters and celebrities. For this purpose, we extract the segments labeled with the same pseudo-speakers based on the tape-level diarization output. These segments that belong to the same speakers are merged to create a single recording for each pseudo-speaker. The same procedure is applied to the speaker segments with known identities. The merged segments that are shorter than a threshold duration are removed to ensure some amount of presence of the pseudo-speakers on tape level. 

For all remaining pseudo-speakers and known speakers, we extract i-vectors and x-vectors using the extractors trained using the implementations available in the Kaldi toolkit~\cite{kaldi}. The performance of these systems is found to be comparable with a similar system trained on the automatically labeled Frisian-Dutch data described in~\cite{yilmaz2017_3}. A similarity matrix for all possible speaker pairs are calculated using a PLDA model~\cite{prince2007} to find the cross-tape speaker similarities. Based on these speaker similarity scores, speaker linking is performed by applying complete-linkage clustering as described in~\cite{ghaemmaghami2016}. Varying levels of speaker linking can be performed by manipulating the clustering threshold results in different levels of speaker and cluster impurities. All speaker labels in the same cluster are mapped to a new pseudo-speaker label or, if a known speaker exists in the cluster, identified as the known speaker.
\vspace{-0.25cm}
\section{Experimental setup}
\label{sec:exps}
\vspace{-0.1cm}
\subsection{Speech Corpora}
\label{ssec:data}

The GMM-UBM model and the i-vector extractor is trained using the following corpora: Switchboard Phase 1-2-3, Switchboard Cellular 1-2, SRE2004-SRE2010 and Mixer 6. The x-vector extractor training has been done using the same corpora and some additional recordings that are noisy and reverberant versions of a randomly selected subset of the clean utterances. The PLDA training for the i-vector and x-vector system has been done on a subset of the augmented data that contains recordings from previous SRE corpora only. The preparation of the FAME! SD corpus on which the speaker linking results are reported is detailed in Section~\ref{sec:database}. 
\vspace{-0.2cm}
\subsection{Implementation details}
\label{ssec:impdet}
We used the LIUM toolkit (ver.~8.4.1) for the tape-level SD in the first stage. This system provides a time averaged tape-level diarization error rate of 19.6\% on all 82 tapes. There are 338 pseudo-speakers in total, while the actual speaker number is 215. The merged pseudospeaker recordings shorter than 10 seconds are not included in the speaker linking stage. The total number of pseudo-speakers identified after the tape level SD for the annotated 82 and all tapes are 811 and 45\,288 respectively. Segments from 187 known speakers are also included during the speaker linking in both linking scenarios. These segments are extracted from the remaining 122 manually annotated segments in the FAME! speech corpus which are excluded from the development and test set.  

The i-vector extractor is trained according to standard Kaldi (v5.4) recipe (\textit{sre16/v1}) developed for the NIST SRE 2016 evaluation. The incorporated GMM-UBM baseline is detailed in \cite{snyder2015}. The acoustic features are 20 MFCCs with a frame shift of 10 ms and a frame length of 25 ms with deltas and delta-deltas. Mean normalization is applied over a 3 second window. For the GMM-UBM, a full-covariance matrix is trained initially by applying 4 EM iterations followed by another 4 iterations with a full-covariance matrix. The i-vector extractor is obtained after 5 EM iteration on the training data and it generates 600-dimensional i-vectors. Finally, the i-vector mean subtraction and length normalization is applied before calculating the PLDA scores. 

\begin{figure*}[!t]
\centering
\begin{subfigure}{.5\textwidth}
  \centering
  \includegraphics[trim=1cm -2.25cm 1cm 2.25cm,width=0.65\linewidth]{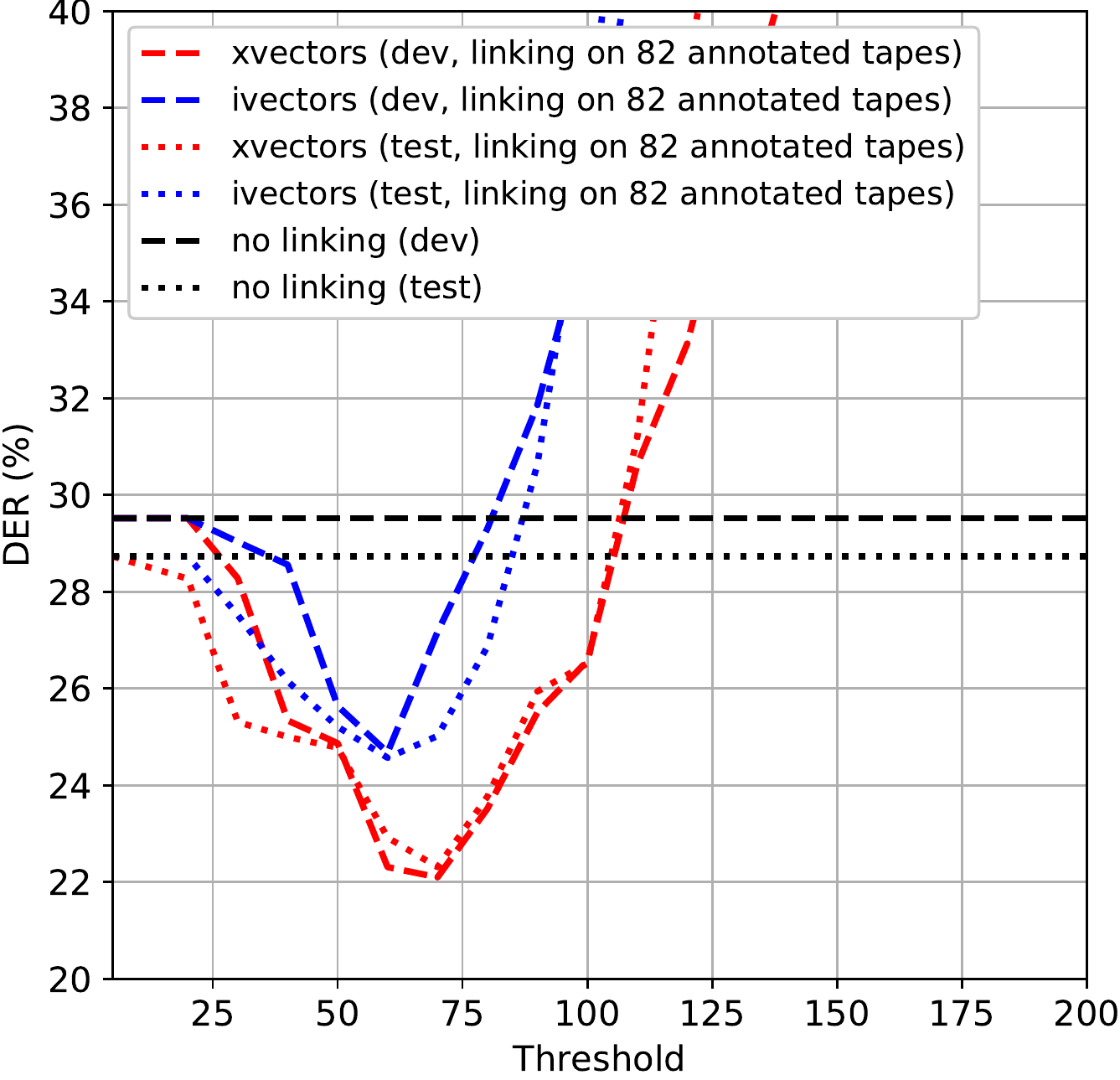}
  \vspace{-1.25cm}
  \caption{DER results - small set}
  \label{fig:sub1}
\end{subfigure}%
\begin{subfigure}{.5\textwidth}
  \centering
  \includegraphics[trim=1cm -2.25cm 1cm 2.25cm,width=0.65\linewidth]{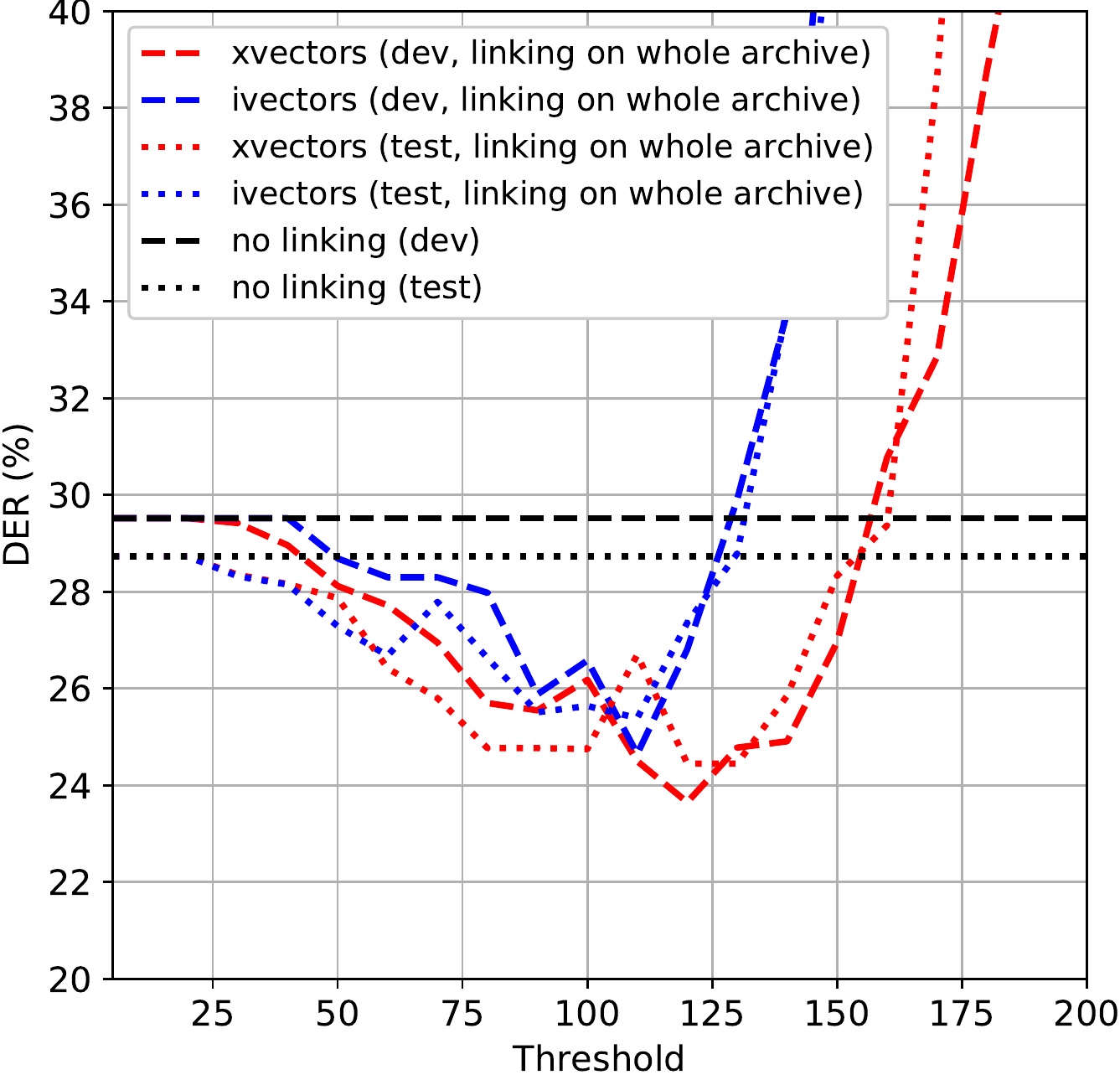}
  \vspace{-1.25cm}
  \caption{DER results - complete set}
  \label{fig:sub2}
\end{subfigure}
\begin{subfigure}{.5\textwidth}
  \centering
  \includegraphics[trim=1cm 0.25cm 1cm -0.25cm,width=0.67\linewidth]{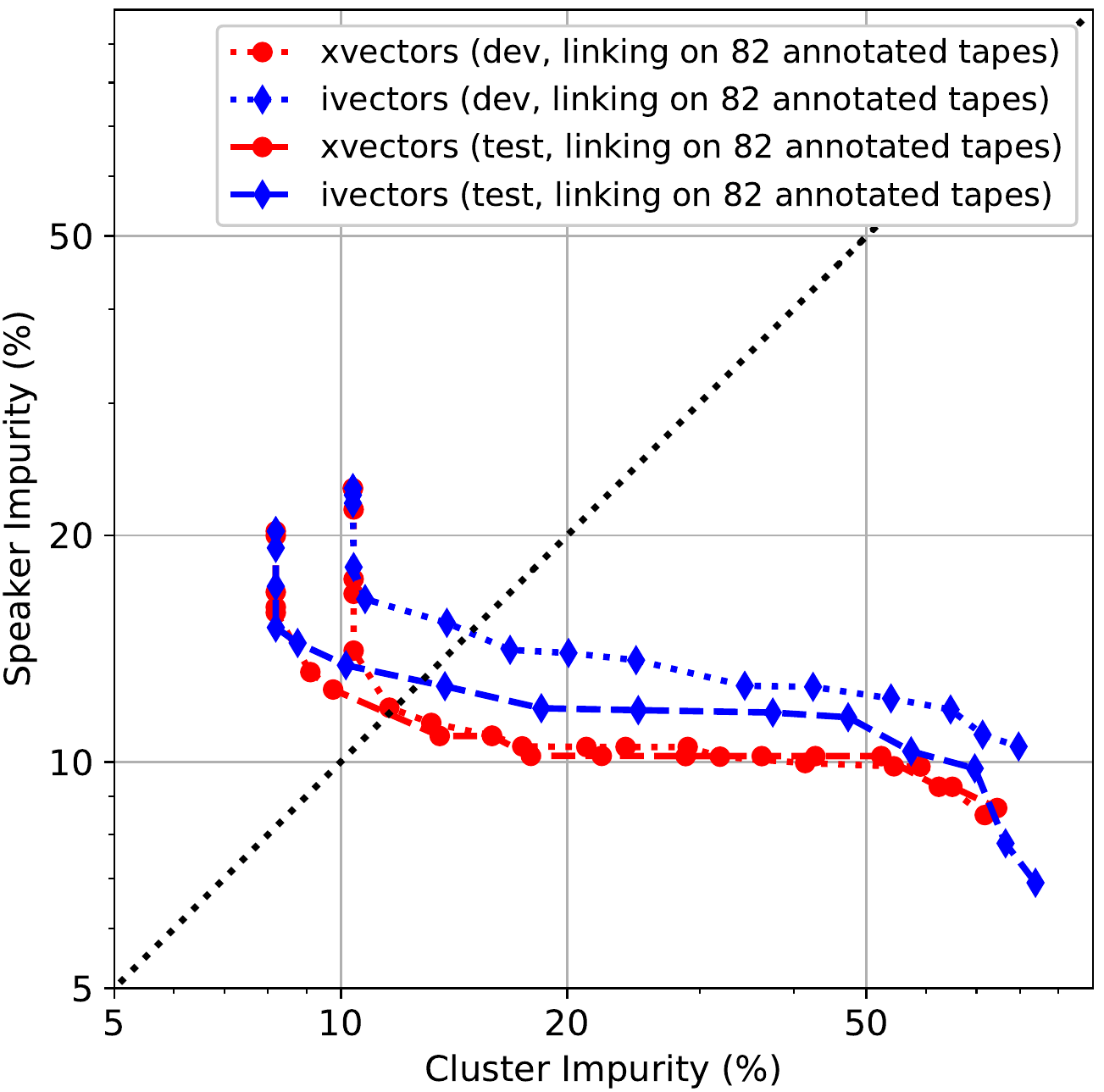}
  \caption{Speaker and cluster impurities - small set}
  \label{fig:sub3}
\end{subfigure}%
\begin{subfigure}{.5\textwidth}
  \centering
  \includegraphics[trim=1cm 0.25cm 1cm -0.25cm,width=0.67\linewidth]{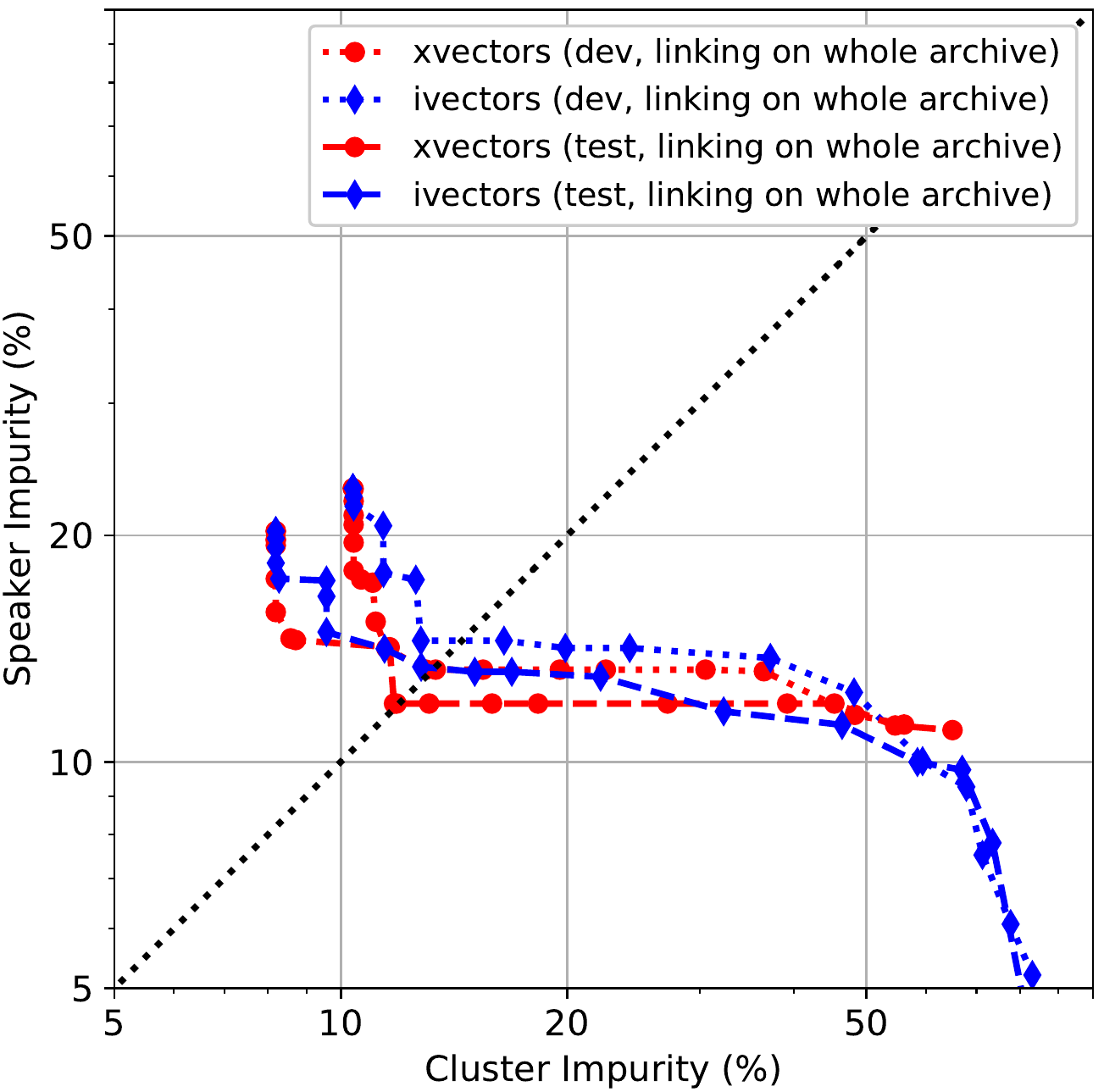}
  \caption{Speaker and cluster impurities - complete set}
  \label{fig:sub4}
\end{subfigure}
\vspace{-0.2cm}
\caption{Speaker linking results obtained on the development and test set}
\vspace{-0.55cm}
\label{fig:der}
\end{figure*}
The x-vector extractor used in these experiments~\cite{snyder2018} is trained according to the standard Kaldi recipe (\textit{sre16/v2}). A time-delay neural network (TDNN)~\cite{waibel1989} is trained using the same acoustic features as in the i-vector system. The TDNN model includes five frame-level hidden layers with rectified linear unit activation and batch normalization~\cite{ioffe2015}. The specific time-delay information of these frame-level layers are given in~\cite{snyder2018}. A statistics pooling layer follows the output of the last frame-level layer which computes the mean and standard deviation of the frames of input segments. The final two hidden layers are 512-dimensional pooling layers, also operating at segment level, prior to the softmax layer which targets speaker labels for each audio segment. The softmax and the second pooling layer are removed during x-vector extraction and 512-dimensional x-vectors are extracted at the output of the first pooling layer. The PLDA estimation has been performed using the clean and augmented SRE data after applying 10 expectation-maximization iterations for both systems.

The SciPy~\cite{scipy} implementation of agglomerative clustering with complete-linkage has been used for performing the speaker clustering. By varying the threshold, various clustering levels have been created and the linking performance for all thresholds has been reported. In the final stage, a common speaker label is assign to the pseudo-speakers and known speakers that belong to the same cluster.
\vspace{-0.2cm}
\subsection{Speaker linking experiments}
\label{ssec:slexps}
We compare the speaker linking performance of i-vectors and x-vectors in the pipeline shown in Figure~\ref{fig:overview} on (1) 82 tapes which contains the annotated subsegments (small set) and (2) all 6494 tapes (complete set). The former scenario contains 53 hours speech data while the latter has approximately 3000 hours of speech data. We evaluate the speaker linking performance using two metrics: (1) diarization error rate (DER) calculated by concatenating the 41 annotated segments in the development and test set and (2) speaker and cluster impurities~\cite{vanleeuwen2010} obtained on the development and test data. The former metric measures the speaker confusions between different clusters, the latter measures the amount of different speakers in each cluster.
\vspace{-0.2cm}
\section{Results and Discussion}
\label{sec:res}
The DER results obtained on each set are presented in Figure~\ref{fig:sub1} and \ref{fig:sub2} for the small and complete linking scenario respectively.  The DER value without performing any speaker linking is 29.5\% (28.7\%) on the development (test) set as shown in black dashed (dotted) line. By varying the clustering threshold, we investigate where each system reaches the lowest DER for linking on the small and complete set. For the linking on small set, the i-vector system has a lowest DER of 24.7\% (24.6\%), while the x-vector system gives a DER of 22.1\% (22.3\%). Both results are much lower than the DER value without linking indicating that the described pipelines improve the quality of the speaker labels by linking the speakers appearing in multiple tapes. Moreover, the x-vector system outperforms the i-vector system with a large margin on both sets. The best results are achieved at a clustering threshold of 60 for the i-vector system and 70 for the x-vector system on both sets. 

The lowest DER values increase for both systems when linking is done on the complete archive. The DER values for the i-vector and x-vector systems are 24.7\% (25.4\%) and 23.7\% (24.5\%) respectively. All results are worse than the previous results reported when the linking is done on the small set. Scaling up from 82 to 6494 tapes increases the complexity of the linking task considerably resulting in some linking accuracy loss in general. The improvements given by the x-vectors over i-vectors are smaller in this case compared to the small linking scenario. The scale up in the amount of tapes included during linking reduces the performance of the x-vector system more dramatically than the i-vector system. The thresholds providing the lowest DER are larger than the corresponding thresholds in the small linking scenario. For a specific system, the thresholds providing the lowest DERs for development and test sets coincide in both small and complete linking scenarios implying that a threshold chosen on the development set generalizes well to the unseen test data.

The speaker and cluster impurities obtained for varying thresholds are plotted in Figure~\ref{fig:sub3} and \ref{fig:sub4} for the small and complete linking scenario respectively. Similar to the DER results, the x-vector system provides the lowest equal impurity value with 11.7\% (11.6\%) on the small linking set, while the i-vector system has an equal impurity of 14.8\% (12.8\%). Scaling up to the complete archive increases the equal impurity on both sets for all systems, except the i-vector system which gives comparable results on the development set. The x-vector system performs better on both sets compared to the i-vector system with smaller improvements than the ones reported on the small linking scenario as also observed in the DER results. In general, it can be concluded from both the DER and impurity results that using x-vectors for speaker linking provides better results than the i-vector system which is consistent with earlier literature on other speaker-related applications. On the other hand, the performance losses reported due to the scaling up in the amount of linked tapes are consistently larger for the x-vector system.
\vspace{-0.25cm}
\section{Conclusion}
\label{sec:conc}
\vspace{-0.15cm}
In this paper, we present a new corpus consisting of 6500 tapes from a bilingual radio archive designed for large-scale speaker diarization research and investigate the speaker linking performance of a new x-vector-based linking system by performing speaker clustering using a similarity matrix with PLDA scores. The clustering is achieved by performing agglomerative clustering with complete-linkage and performed using speaker models from all pseudo-speakers labeled by a generic SD system and a small set of known speakers. Based on the discovered clusters, we perform speaker linking and identification by detecting similarities between the pseudo-speakers appearing in different tapes and identify them if they belong to the same cluster as a known speaker. We compare the speaker linking performance of i-vector and x-vector variants of the described pipeline on the new SD corpus in terms of average diarization error rates and speaker and cluster impurities. Future work includes exploring the recently proposed meta-embeddings~\cite{brummer2018} for speaker clustering using the introduced experimental setup.
\vspace{-0.25cm}
\section{Acknowledgements}
\vspace{-0.15cm}

This research is supported by National Research Foundation through the AI Singapore Programme, the AI Speech Lab: Automatic Speech Recognition for Public Service Project AISG-100E-2018-006 and the NWO Project 314-99-119.
\bibliographystyle{IEEEtran}
\balance
\bibliography{refs}

\end{document}